\documentclass[pdflatex,sn-nature]{sn-jnl}

\usepackage{graphicx}%
\usepackage{multirow}%
\usepackage{amsmath,amssymb,amsfonts}%
\usepackage{amsthm}%
\usepackage{mathrsfs}%
\usepackage[title]{appendix}%
\usepackage{xcolor}%
\usepackage{algorithm}%
\usepackage{listings}%
\usepackage{times}
\usepackage{soul}
\usepackage{url}
\usepackage[utf8]{inputenc}
\usepackage[small]{caption}
\usepackage{amsmath}
\usepackage{booktabs}
\usepackage{algorithmic}
\usepackage[switch]{lineno}
\usepackage[T1]{fontenc}
\usepackage{enumitem}
\usepackage{multirow}
\usepackage{makecell}
\usepackage{subcaption}
\usepackage{tabularx}
\usepackage[most]{tcolorbox}
\usepackage{makecell} 
\usepackage{tikz}
\usepackage{verbatim}
\usepackage{soul}
\usepackage{adjustbox}
\usepackage{float}
\tcbset{colback=gray!5!white, colframe=gray!60!blue, coltitle=white, boxrule=0.5pt, arc=2mm, left=4pt, right=4pt, top=4pt, bottom=4pt, fonttitle=\bfseries}

\usepackage{thmtools}
\usepackage{thm-restate}
\usepackage{setspace}

\definecolor{personaheader}{RGB}{75, 68, 128}
\definecolor{personabg}{RGB}{235, 238, 245}

\definecolor{evidenceheader}{RGB}{100, 130, 85}
\definecolor{evidencebg}{RGB}{235, 245, 232}

\theoremstyle{thmstyleone}%
\theoremstyle{thmstyletwo}%

\theoremstyle{thmstylethree}%
\newtheorem{definition}{Definition}%

\begin{document}

\title[]{Shaping AI Explanations to Expert Perspectives for Scientific Discovery}

%

%
%
%

\author*[]{\fnm{Susana} \sur{Nunes}}\email{scnunes@ciencias.ulisboa.pt}

\author[]{\fnm{Tiago} \sur{Guerreiro}}\email{tjguerreiro@ciencias.ulisboa.pt}

\author[]{\fnm{Catia} \sur{Pesquita}}\email{clpesquita@ciencias.ulisboa.pt}

\affil[]{LASIGE, Faculdade de Ciências da Universidade de Lisboa, Portugal}

\abstract{
AI explanations for scientific discovery are typically optimized around a single notion of explanation quality. Experts, however, differ in how they assess evidence, prioritize mechanisms, and judge relevance, meaning that a general-purpose explanation may not provide the evidence an expert finds most informative or convincing. We show that these differences exhibit structured epistemic perspectives and ask whether explanations conditioned on these perspectives are preferred by experts. To study this, we introduce a reinforcement learning-based framework for generating \textit{perspective-conditioned explanations}, which is guided by the judgments of LLM-based personas derived from expert evaluations.
We instantiate the framework over scientific knowledge graphs and evaluate it in drug discovery through drug repurposing and drug--target interaction prediction. Expert studies show that perspective-conditioned explanations capture characteristic patterns of judgment, are preferred over general-purpose explanations, and receive higher relevance and validity ratings without sacrificing predictive performance. These findings establish explanation quality as perspective-dependent and enable scalable, perspective-aligned explanations for scientific discovery.}




\maketitle




Scientific discovery increasingly relies on AI-generated predictions to guide hypothesis formulation, evaluate candidate mechanisms, and prioritize experimental targets~\cite{wang2023scientific}. However, prediction alone is insufficient: scientists rely on interpretable outputs to understand model reasoning, evaluate alignment with established domain knowledge, and integrate evidence into the iterative cycle of theory refinement~\cite{miller2019explanation}. Explainable AI (XAI) seeks to support this process by exposing the basis for model outputs, but explanation quality is typically treated as a property of the explanation itself. In scientific reasoning, this assumption is limiting. Experts can examine the same evidence and differ in which mechanisms they prioritize, which explanatory alternatives they find informative, how they weigh uncertainty and prior knowledge, and ultimately which explanation they consider most convincing.

We refer to these structured differences in the evaluation of evidence as \textit{epistemic perspectives}. Existing adaptive, personalized, interactive, and human-centered XAI approaches recognize that explanations need not be identical for all users, but typically condition adaptation on characteristics such as professional role, expertise level, or interaction history~\cite{delaunay2023adaptation,conati2021toward,slack2023explaining}. These characteristics describe \textit{who} the user is, but do not necessarily capture \textit{how} that user evaluates scientific evidence. This distinction is particularly important in scientific domains, where experts differ systematically in their preference for mechanistic versus associative evidence~\cite{jimenez2020drug,wu2023black}, their requirements for scientific validity~\cite{ponzoni2023explainable}, and how they weigh uncertainty and prior knowledge~\cite{schaffer2019can,gaube2021ai}. Explanation quality may therefore be relational rather than intrinsic, depending not only on the explanatory evidence itself but also on the epistemic perspective through which that evidence is interpreted. This motivates \textit{perspective-conditioned explanations}: explanations whose evidential content and structure are shaped by differences in how experts judge scientific evidence.

Studying this form of perspective dependence requires explanations whose evidential structure can be explicitly inspected and varied. Scientific Knowledge Graphs (KGs) are well suited to this setting because they represent domain knowledge as structured, multi-relational networks of entities and relations~\cite{lecue2020role,tiddi2022knowledge,rajabi2024knowledge}. Predictions can then be supported by explicit relational paths that connect the entities of a scientific hypothesis and expose the evidence and mechanisms underlying that connection. Reinforcement Learning (RL) approaches operationalize this process by treating path generation as a sequential decision problem over the graph~\cite{xiong2017deeppath,das2017go}. Because multiple plausible paths can support the same prediction, KG explanations provide a setting in which different evidential emphases can be expressed through which paths are selected, which mechanisms are foregrounded, and how evidence is combined. Existing KG-based explanation methods, however, generally optimize shared criteria such as fidelity, logical consistency, plausibility, or relevance~\cite{nunesRewarding2025,huang2024foundation,perdomo2026generating}. They therefore assume a generic evaluator rather than accounting for systematic differences in how experts assess the same explanatory evidence.

In this work, we ask whether such epistemic variation can be identified, modeled, and used to shape explanation generation. We first collect expert preferences and judgments over KG-based AI explanations in drug discovery. Clustering these preferences reveals distinct and interpretable groups corresponding to different epistemic perspectives. Importantly, these groups are not imposed \textit{a priori}, but emerge from expert judgments, providing evidence that epistemic variation is structured and measurable rather than merely idiosyncratic disagreement. This observation motivates our approach: if recurring patterns exist in how experts evaluate explanations, they can potentially provide a signal for explanation generation.

Directly using expert judgments to optimize explanation generation, however, is difficult to scale. Evaluating a scientific explanation is a multidimensional task that requires careful inspection of its mechanistic and evidential content. In our setting, experts must assess multi-hop explanatory paths for relevance, validity, and completeness, a substantially richer task than providing simple preference labels. Large-scale direct expert supervision is therefore impractical, motivating a representation of the shared evaluative structure revealed by expert judgments. We capture this structure through \textit{epistemic personas}, computational proxies for recurring epistemic perspectives. Rather than representing users through demographic or professional attributes, these personas encode \textit{how} experts evaluate explanations, including how they balance mechanistic detail, explanatory complexity, evidential relevance, and scientific validity. Because they are grounded in observed expert judgments rather than generated \textit{a priori}, they allow the patterns of expert evaluation to be applied beyond the individual experts from whom they were derived.

\begin{figure*}[tb!]
\centering
\includegraphics[width=\linewidth]{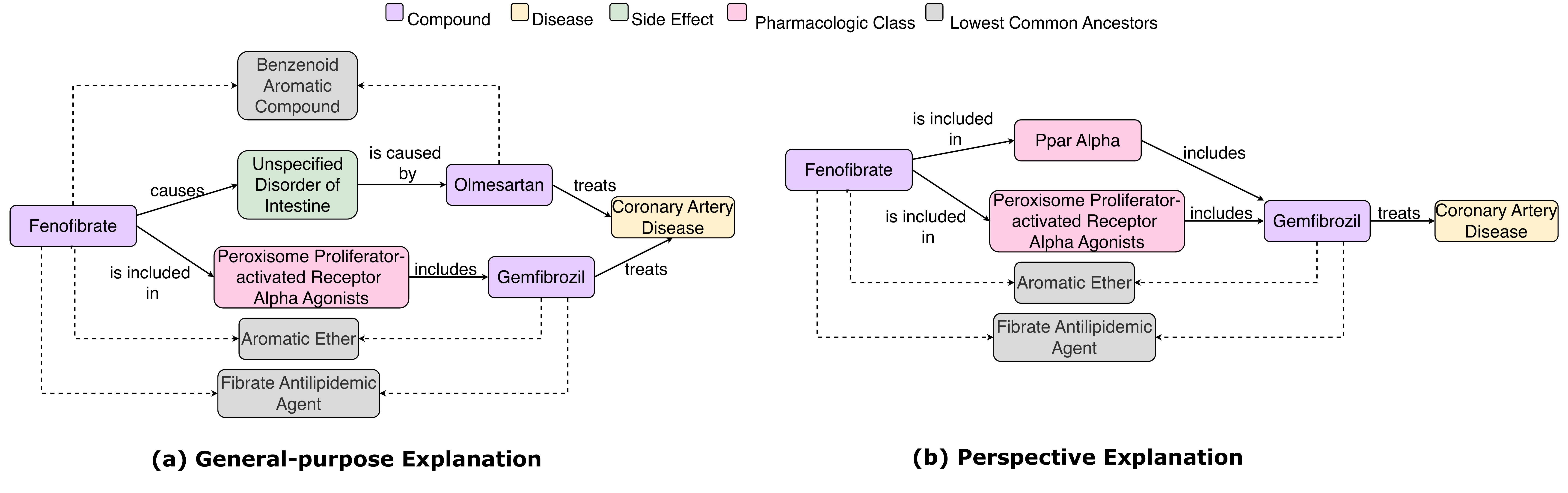}
\caption{General-purpose and perspective-conditioned explanations for \textit{Fenofibrate treats Coronary Artery Disease}. (a) The general-purpose explanation generated by REx includes peripheral paths such as side effects and loosely related drug classes. (b) The perspective-conditioned explanation, aligned with a perspective that prioritizes mechanistic directness, selects paths focused on the core therapeutic mechanism and relevant biological targets.}
\label{fig:exp-pair}
\end{figure*}

We incorporate these perspectives directly into explanation generation by formalizing perspective conditioning as a reinforcement learning problem over explanatory paths. Epistemic personas evaluate candidate paths according to their epistemic perspective, and their judgments provide reward signals that guide the RL agent toward different selections and combinations of evidence. Perspective conditioning therefore changes the optimization objective itself, rather than adapting the presentation of an explanation only after it has been generated. Figure~\ref{fig:exp-pair} illustrates this distinction. For the same prediction, the general-purpose explanation includes peripheral evidence such as side effects and loosely related drug classes, whereas the perspective-conditioned explanation foregrounds paths that directly support the therapeutic mechanism and relevant biological targets. The difference is not simply one of explanation length or level of detail, but of which evidence is considered explanatory under a particular epistemic perspective.

We evaluate the framework in drug discovery using drug repurposing and drug-target interaction prediction as complementary tasks. Drug discovery provides a demanding setting in which experts must reason across heterogeneous molecular, mechanistic, and clinical evidence, while the two tasks allow us to examine whether recurring evaluative patterns extend across different scientific prediction problems. Across expert studies, the inferred personas capture characteristic patterns of expert judgment, and perspective-conditioned explanations are preferred over general-purpose explanations in 63.3--76.0\% of comparisons and receive higher ratings for relevance and validity while preserving predictive performance. Persona-based evaluation also reduces the amount of direct expert feedback required for optimization by approximately two orders of magnitude.

These findings show that variation in expert judgments of scientific explanations contains structured information about how experts evaluate evidence, and can be made actionable during explanation generation. More broadly, our results challenge the assumption that there is a single generally optimal explanation for a model prediction. Instead, they support a view of explanation quality as perspective-dependent and show how modeling epistemic variation can enable scalable, perspective-aligned explanation generation for scientific discovery.

\section*{Results}

\begin{figure*}[htb!]
    \centering
    \includegraphics[width=\linewidth]{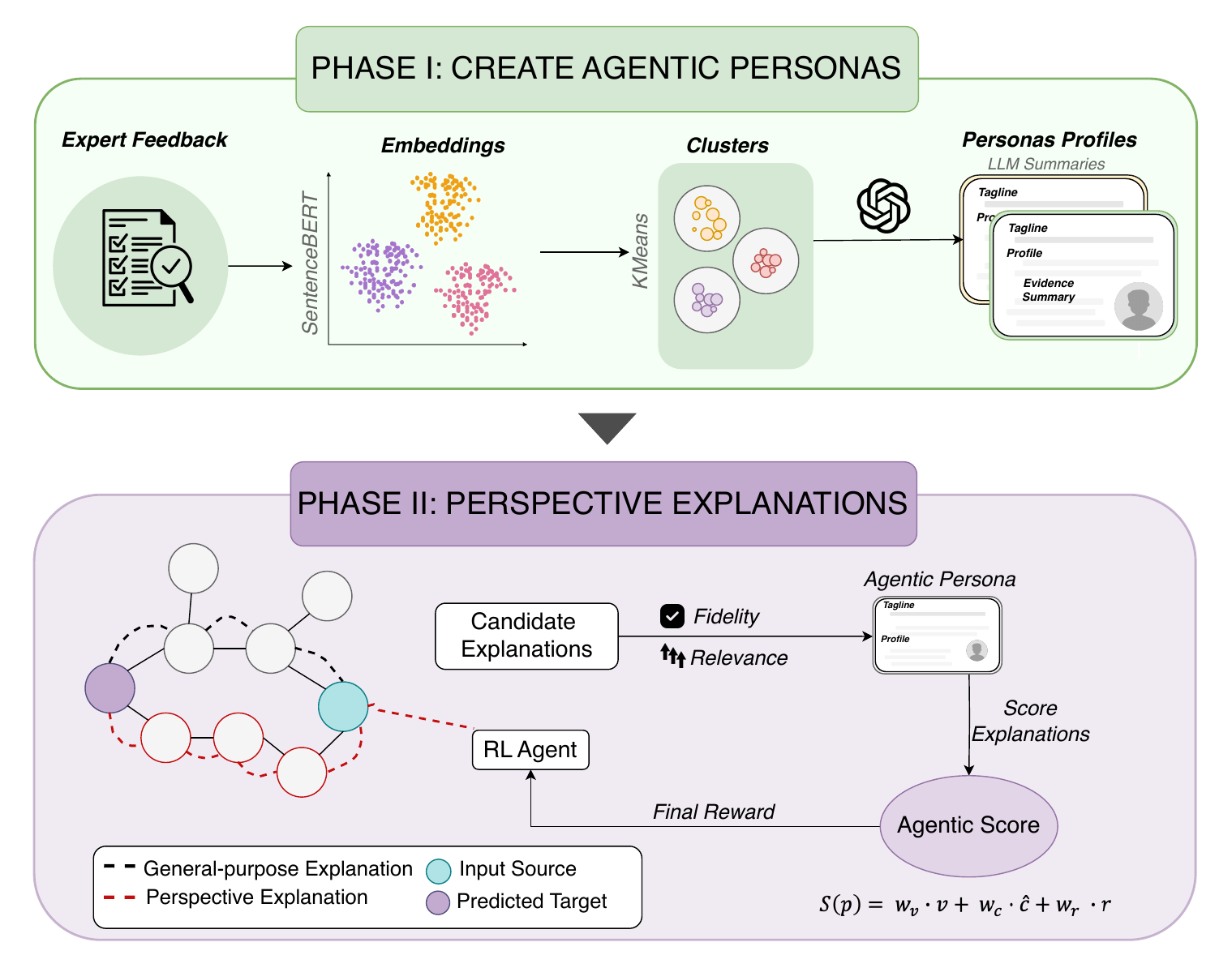}
    \caption{Overview of the perspective explainability approach. Phase~I creates epistemic personas by embedding expert qualitative feedback, clustering responses to identify shared evaluative patterns, and synthesizing structured persona narratives via LLM summarization. Phase~II integrates these personas into the reinforcement learning loop: as the RL agent generates candidate explanation paths over the knowledge graph, each persona scores them according to its epistemic perspective, producing a persona-conditioned reward signal that guides the agent toward explanations aligned with distinct expert reasoning strategies.}
    \label{fig:overview}
\end{figure*}

Building on prior work in knowledge graph–based explanation generation via reinforcement learning~\cite{nunesRewarding2025}, we reframe the explanation process as epistemically adaptive reasoning. In the original formulation, an RL agent traverses a scientific knowledge graph to discover explanatory paths connecting entities present in a hypothesis statement. For example, in a drug repurposing task, given the hypothesis that \textit{Fenofibrate treats Coronary Artery Disease}, the agent uncovers relevant paths between the two entities (see Figure~\ref{fig:exp-pair}a).  

We transform this paradigm into a two-phase, persona-conditioned framework (Figure~\ref{fig:overview}) in which path discovery and explanation framing are explicitly guided by epistemic personas and a persona-aligned reward function. Rather than generating generic explanations, the agent now selects and structures knowledge graph paths in accordance with distinct epistemic perspectives, enabling the adaptation of explanatory emphasis, such as prioritizing mechanistic explanations (see Figure~\ref{fig:exp-pair}b). 

In Phase~I, we derive epistemic personas from expert feedback: qualitative evaluations of knowledge graph-based explanations produced by domain experts are semantically embedded, clustered to identify shared evaluative patterns, and synthesized into structured persona narratives via LLM-based summarization. In Phase~II, inspired by RLHF~\cite{christiano2017deep}, each persona is instantiated as an LLM prompted with its narrative profile and used to score the candidate explanation paths generated by the RL agent, replacing direct human feedback with a persona-conditioned reward signal that steers the agent toward explanations aligned with that persona's reasoning preferences.

Drug development naturally supports the evaluation of perspective-conditioned explanations because it requires scientists to interpret heterogeneous evidence through diverse perspectives. In tasks such as drug repurposing and drug-target interaction prediction, experts must consider mechanistic hypotheses, molecular associations, pathway-level evidence, prior biological knowledge, and potential clinical relevance. These tasks therefore invite different interpretive stances: some experts may prioritize mechanistic specificity, while others emphasize evidential robustness or translational relevance. We evaluate our framework on these two tasks using Hetionet~\cite{himmelstein2017systematic}, a heterogeneous biomedical knowledge graph that integrates multiple entity and relation types relevant to drug development, including drugs, diseases, genes, pathways, side effects, and biological processes. This heterogeneity makes Hetionet well-suited for studying perspective-conditioned explanation generation: the same prediction can be supported by multiple plausible explanatory paths, allowing epistemic perspectives to be expressed through which paths are selected, how evidence is composed, and what forms of biomedical evidence are emphasized. 

We derive epistemic personas from a persona creation study ($n=11$) and assess the effectiveness of persona-conditioned perspective explanations in a persona evaluation study ($n = 22$). Given the cost and cognitive burden of expert evaluation of structured explanations, our goal is not large-scale population coverage, but the identification of recurring and internally consistent evaluative structure within a particular expert cohort. In expert-driven scientific domains, increasing sample size beyond this scale is often infeasible due to the scarcity of qualified evaluators, the depth of per-instance annotation required, and the iterative cognitive effort required to judge scientific explanations. In fact, this constraint also motivates our use of epistemic personas within the reinforcement learning framework, as a mechanism for scaling evaluative feedback beyond direct expert supervision.

We next evaluate whether expert feedback contains recoverable epistemic structure, whether this structure can be represented through epistemic personas, and whether persona-conditioned rewards can steer explanation generation toward outputs that experts judge as more relevant and scientifically valid.

\subsection*{Experts differ in how they evaluate explanations}
\label{sec:clustering_results}
To identify systematic differences in how experts interpret explanations, we conducted a persona creation study with 11 biomedical experts spanning life sciences, computer science/AI with biomedical experience, and hybrid computational biology backgrounds (demographics in Supplementary Table~\ref{tab:supp-formative}). Participants assessed knowledge graph-based explanations for 10 drug repurposing (DR) hypotheses and 10 drug-target interaction (DTI) hypotheses. For each hypothesis, explanations were generated by four representative systems with distinct explanation strategies: MINERVA~\cite{das2017go}, which optimizes for path fidelity, i.e., whether the path correctly connects the subject and object; PoLo~\cite{liu2021neural}, which adds logical-rule constraints to the reward; REx~\cite{nunesRewarding2025}, which jointly optimizes fidelity and relevance and enriches paths with ontological context; and RExLight, an ablation of REx without ontological enrichment.

Participants evaluated each explanation using both detailed qualitative feedback and quantitative ratings along three dimensions derived from established frameworks of explanatory virtues in scientific theories~\cite{keas2018systematizing,nunes2026ESWC}: \textit{relevance}, \textit{completeness}, and \textit{validity}. \textit{Relevance} captures whether an explanation provides informative causal insight into the prediction beyond generic relational structure. \textit{Completeness} reflects the extent to which the explanation provides sufficient causal detail to support the prediction while maintaining parsimony, balancing richness against redundancy. \textit{Validity} assesses whether the proposed causal mechanism is biologically plausible and consistent with established biomedical knowledge~\cite{rosales2021scientific}, including whether intermediate steps correspond to known scientific processes rather than spurious associations. All three dimensions were rated using 5-point Likert scales.

Qualitative feedback was provided by a subset of participants: of the nine who contributed, six evaluated both DR and DTI, and three evaluated DR only, yielding 15 task-level responses. Across the 40 explanations evaluated per task, expert ratings varied substantially, indicating systematic differences in how experts assess the quality of knowledge graph-based scientific explanations.

As shown in Figure~\ref{fig:persona-discovery}a and \ref{fig:persona-discovery}b, ratings exhibit high variance (SD 0.9--1.4) even for top-performing systems, underscoring the interpretive diversity across experts that personas aim to capture. Nevertheless, there is a broad consensus among experts, with more sophisticated systems consistently receiving higher ratings.

\subsection*{Two distinct epistemic perspectives emerge from expert feedback}

\begin{figure}[H]
\includegraphics[width=1\linewidth]{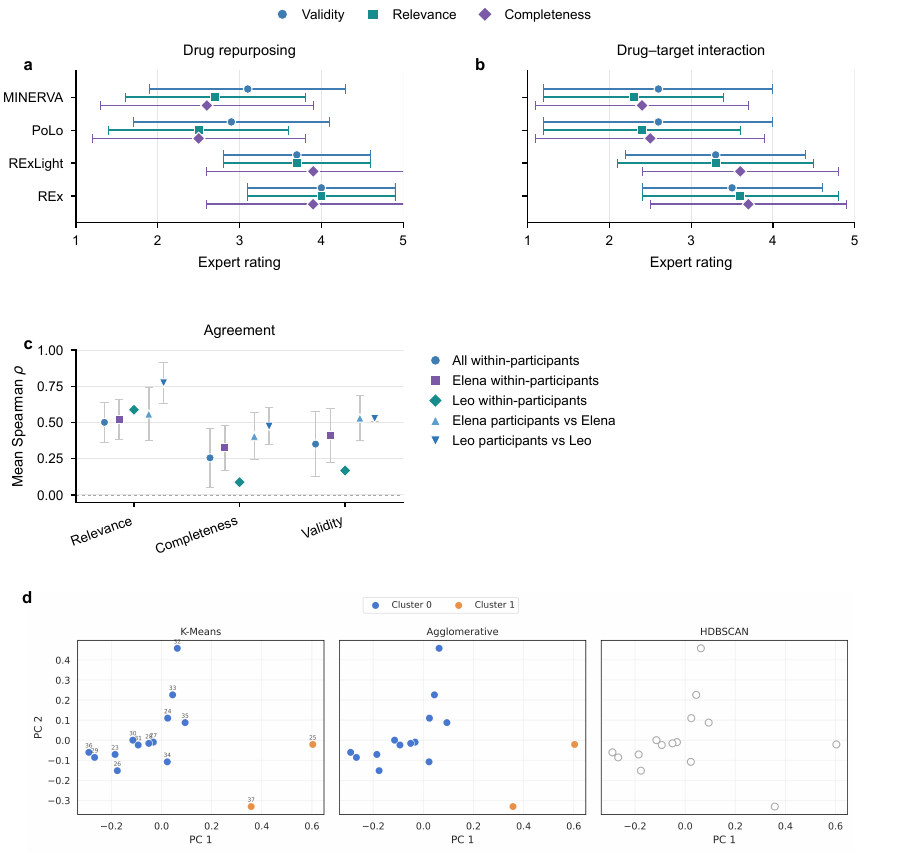}
\caption{Persona discovery: quantitative evidence. \textbf{a,b}, Persona creation ratings by system and task (means $\pm$ SD on 1--5 scale), showing high variance reflecting interpretive diversity. \textbf{c}, Mean pairwise Spearman correlations ($r$) between expert ratings and between expert and persona ratings across all drug repurposing and drug-target interaction tasks; all $p<.001$. \textbf{d}, Clustering of expert feedback embeddings projected via PCA, comparing K-Means, Agglomerative, and HDBSCAN algorithms. K-Means and Agglomerative converge on an identical two-cluster solution, while HDBSCAN does not resolve cluster structure.}
\label{fig:persona-discovery}
\end{figure}

To identify groups with similar epistemic perspectives, we embedded participants' qualitative feedback using Sentence-BERT~\cite{reimers-2019-sentence-bert} and applied unsupervised clustering. K-Means and Agglomerative clustering converged on an identical $k=2$ solution (100\% agreement), while HDBSCAN did not resolve cluster structure (Figure \ref{fig:persona-discovery}d). Among candidate solutions ($k=2$–$5$), $k=2$ achieved the highest silhouette score (0.29) and Calinski–Harabasz index. Gaussian mixture model selection independently confirmed the two-component solution, with $k=2$ achieving the lowest Bayesian information criterion (BIC) and Akaike information criterion (AIC). Bootstrap resampling yielded the highest cluster stability at $k=2$ (Jaccard = 0.74 $\pm$ 0.20), substantially exceeding $k=3$ (0.46) and $k=4$ (0.50) (Supplementary Figure \ref{fig:clustering-validation}a).

Mixture and stability checks revealed no consistent subclusters: splitting the 13-responses majority cluster via K-Means yielded a silhouette of only 0.13, and HDBSCAN fragmented it into three unstable micro-groups with six noise points, and per-sample silhouette analysis confirmed that all participants had positive coefficients under the $k=2$ solution (Supplementary Figure \ref{fig:clustering-validation}b). All participants were directly assigned to clusters with no singletons. The resulting clusters were named Elena (13 responses) and Leo (2 responses) for their respective personas.

\begin{figure}[]
\centering
\begin{minipage}[t]{0.48\linewidth}
{\sffamily\textbf{a}}\\
\vspace{0pt}
\begin{tcolorbox}[
  colback=personabg, colframe=personaheader, colbacktitle=personaheader,
  coltitle=white, title={\bfseries \footnotesize Elena Persona Narrative},
  fonttitle=\sffamily\tiny, fontupper=\sffamily,
  arc=2mm, boxrule=0pt, toptitle=0.5mm, bottomtitle=0.5mm,
  left=1.5mm, right=1.5mm, top=0.5mm, bottom=0.5mm,
  width=\linewidth
]
\setlength{\baselineskip}{0.95em}
\begin{footnotesize}\raggedright\setlength{\parskip}{0pt}%
\textbf{Tagline:} \textit{Trusts nothing but rock-solid mechanistic logic.}\\
\textbf{Profile:} Elena weighs every therapeutic claim against hard mechanistic evidence. She expects an explanation to start with a clinically proven drug--disease link or a well-documented mode of action: microtubule inhibition, T-cell suppression, a binding $IC_{50}$ in hand. Side-effect coincidences, vague ``causes'' predicates, or generic class labels make her skeptical unless an ontological note shows exactly why that class matters biologically. She welcomes extra layers: genetic markers, adverse-effect profiles, ontology tags, only when they tie directly to the core mechanism and keep the path lean. Superficial complexity or redundant paths lose her attention; a handful of well-justified edges beats a tangled web every time.\\
\textit{Also values\ldots} Occasional ontological expansions that rescue an otherwise hidden mechanistic clue, provided they do not swamp the narrative.\\
\textit{Background note:} Elena's community is split between bench researchers and AI modelers, and she holds both to the same rigorous standard of scientific plausibility.
\end{footnotesize}
\end{tcolorbox}
\vspace{2pt}
\begin{tcolorbox}[
  colback=evidencebg, colframe=evidenceheader, colbacktitle=evidenceheader,
  coltitle=white, title={\bfseries \footnotesize Elena Evidence Summary},
  fonttitle=\sffamily\tiny, fontupper=\sffamily,
  arc=2mm, boxrule=0pt, toptitle=0.5mm, bottomtitle=0.5mm,
  left=1.5mm, right=1.5mm, top=0.5mm, bottom=0.5mm,
  width=\linewidth
]
\setlength{\baselineskip}{0.95em}
\begin{footnotesize}\raggedright\setlength{\parskip}{0pt}%
\textsc{Core traits} ($\geq$40\% support)
\begin{itemize}[nosep,leftmargin=1em,topsep=0pt]
\item Direct, mechanistically justified drug--disease links demanded by 13 of 13 users (100\%).
\item Precise relation types over vague ``causes'' or side-effect chains preferred by 13 users (100\%).
\item Ontological context accepted only when it sharpens the mechanism (11 users, 85\%).
\item Concise yet context-rich presentations that avoid redundancy (9 users, 69\%).
\end{itemize}
\textsc{Secondary traits} (25--40\% support)
\begin{itemize}[nosep,leftmargin=1em,topsep=0pt]
\item Integration of adverse-effect, genetic, and alternative-therapy context to enrich reasoning (5 users, 38\%).
\end{itemize}
\textsc{Weak signals} ($<$25\% support)
\begin{itemize}[nosep,leftmargin=1em,topsep=0pt]
\item Fixed path length targets (1 user, 8\%). \item Sequence-motif or external database citations (1 user, 8\%).
\end{itemize}
\setlength{\baselineskip}{0.95em}
\textsc{Background:} 5 life-science, 4 hybrid computational biology, 4 CS/AI with biomedical focus.\\
\textsc{Concerns:} Unanimity on mechanistic rigor is strong, but preferences for path length or specific external sources are isolated and may not generalize.
\end{footnotesize}
\end{tcolorbox}
\end{minipage}\hfill
\begin{minipage}[t]{0.48\linewidth}
{\sffamily\textbf{b}}\\
\vspace{0pt}
\begin{tcolorbox}[
  colback=personabg, colframe=personaheader, colbacktitle=personaheader,
  coltitle=white, title={\bfseries \footnotesize Leo Persona Narrative},
  fonttitle=\sffamily\tiny, fontupper=\sffamily,
  arc=2mm, boxrule=0pt, toptitle=0.5mm, bottomtitle=0.5mm,
  left=1.5mm, right=1.5mm, top=0.5mm, bottom=0.5mm,
  width=\linewidth
]
\setlength{\baselineskip}{0.95em}
\begin{footnotesize}\raggedright\setlength{\parskip}{0pt}%
\textbf{Tagline:} \textit{Looks for one illuminating link that makes the mechanism click.}\\
\textbf{Profile:} Leo scans an explanation until a single, definitive connection locks the story into place: an ontological expansion that clarifies a hidden relation or a standout feature like ``dependence on opiates'' that tags the drug's true family. Once that anchor appears, he is satisfied so long as the rest of the path stays short and avoids redundant steps. He is comfortable with a couple of alternative paths for completeness, provided each earns its keep and the display never bloats into an unreadable maze.\\
\textit{Also values\ldots} Using ontology sparingly to light up indirect relations that a simple ``treats'' or ``causes'' link cannot explain.\\
\textit{Background note:} Half of Leo's colleagues come from wet-lab pharmacology, the other half from AI drug-discovery teams, so he stands at the crossroads of formal ontology and data-driven insight.
\end{footnotesize}
\end{tcolorbox}
\vspace{2pt}
\begin{tcolorbox}[
  colback=evidencebg, colframe=evidenceheader, colbacktitle=evidenceheader,
  coltitle=white, title={\bfseries \footnotesize Leo Evidence Summary},
  fonttitle=\sffamily\tiny, fontupper=\sffamily,
  arc=2mm, boxrule=0pt, toptitle=0.5mm, bottomtitle=0.5mm,
  left=1.5mm, right=1.5mm, top=0.5mm, bottom=0.5mm,
  width=\linewidth
]
\setlength{\baselineskip}{0.95em}
\begin{footnotesize}\raggedright\setlength{\parskip}{0pt}%
\textsc{Core traits} ($\geq$40\% support)
\begin{itemize}[nosep,leftmargin=1em,topsep=0pt]
\item Preference for uniquely informative, mechanism-clarifying links (2/2, 100\%).
\end{itemize}
\textsc{Secondary traits} (25--40\%)
\begin{itemize}[nosep,leftmargin=1em,topsep=0pt]
\item Judicious ontological expansion for indirect relations (1/2, 50\%).
\item Limiting path length to avoid complexity (1/2, 50\%).
\end{itemize}
\textsc{Weak signals} ($<$25\%)
\begin{itemize}[nosep,leftmargin=1em,topsep=0pt]
\item None (all mentions $\geq$50\%, small sample size).
\end{itemize}
\textsc{Background:} 1 life-science, 1 CS/AI with biomedical expertise.\\
\textsc{Concerns:} With only 2 contributors, even shared themes rest on minimal evidence.
\end{footnotesize}
\end{tcolorbox}
\vspace{4pt}
{\sffamily\textbf{c}}\\
\vspace{0pt}
\includegraphics[width=\linewidth]{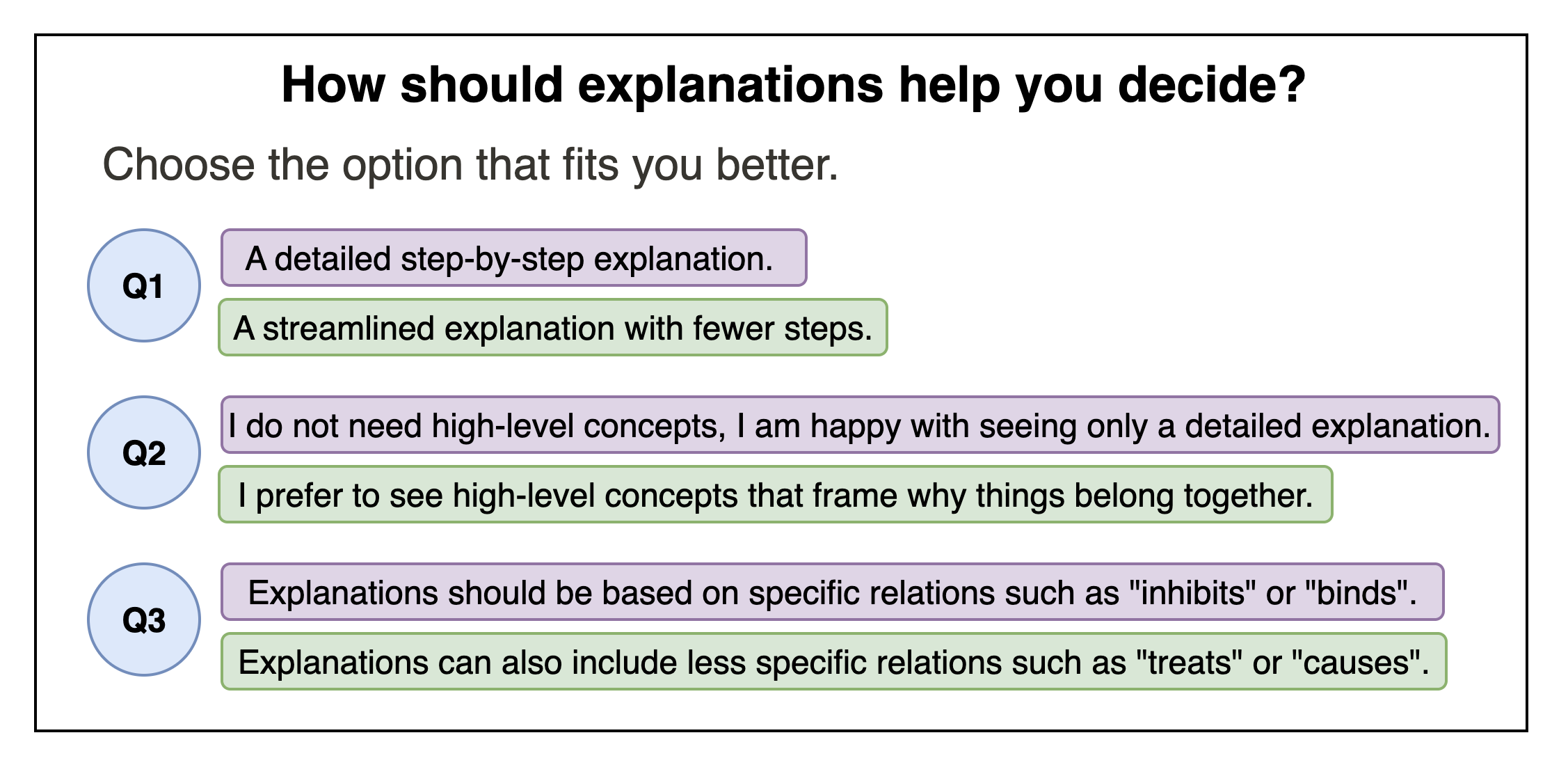}
\end{minipage}
\caption{epistemic persona profiles and assignment. \textbf{a}, Narrative profile and evidence summary for \textit{Elena}, who prioritizes mechanistic rigor. \textbf{b}, Narrative profile and evidence summary for \textit{Leo}, who prioritizes focused clarity. \textbf{c}, Three-question profiling instrument used to assign participants to personas for the persona evaluation study based on discriminative characteristics identified in expert feedback. For each question, the first response option (purple) aligns with Elena's preferences, while the second option (green) aligns with Leo's preferences. Response order was randomized to control for ordering effects.}
\label{fig:persona-profiles}
\end{figure}

To translate clusters into actionable personas, we prompt an instruction-tuned LLM (OpenAI o3-pro) with all user records assigned to each cluster, including curated preference statements and background metadata. The model synthesizes these inputs into structured persona descriptions that represent distinct epistemic perspectives (full narrative profiles and evidence summaries in Figures~\ref{fig:persona-profiles}a and~\ref{fig:persona-profiles}b).

\textit{Elena} reflects mechanistic rigor, demanding biologically grounded, mechanistically explicit explanations and rejecting vague predicates or loosely connected paths. Participants emphasized multiple evidence layers: \textit{``I prefer explanations that combine therapeutic, genetic, adverse-effect, and ontological information for maximal scientific validity.''} Her cluster displayed near-unanimous preferences for clear relation types (e.g., ``inhibits'', ``upregulates'') and ontological layers only when directly relevant.
\textit{Leo} reflects focused clarity, valuing concise relations that highlight a single anchoring connection. Participants called for bounded completeness: \textit{``I prefer explanations that offer multiple paths for completeness while limiting their number to avoid excessive complexity.''} Leo participants penalized over-expansion, emphasizing that explanations should be simple but sufficient.

These personas do not aim to capture the full spectrum of epistemic perspectives in drug discovery; rather, they demonstrate that even within a small expert cohort, distinct evaluative perspectives can emerge from qualitative explanation judgments.

\subsection*{Personas capture structured patterns in how experts evaluate explanation quality}
To assess whether the inferred personas reflect coherent patterns in expert evaluation behavior, we conduct a two-level validation. First, we examine whether experts assigned to the same persona exhibit consistent evaluation strategies when rating the same explanations. Second, we evaluate whether the constructed personas reproduce expert-like evaluation behavior when directly applied to the same set of explanations.

We first analyze within-cluster agreement across validity, relevance, and completeness to test whether the qualitative clustering structure is reflected in consistent quantitative evaluation patterns on the same explanations. We compute mean pairwise Spearman correlations between participants’ ratings within each persona group over the 80 explanations (see Figure~\ref{fig:persona-discovery}c), capturing consistency in relative evaluation behavior. Across all participants, we observe moderate agreement in ranking explanations, with correlations highest for relevance ($\rho = 0.50$), followed by validity ($\rho = 0.35$), and lowest for completeness ($\rho = 0.26$), indicating that experts share a more consistent notion of relevance than of completeness or validity. Stratifying by persona reveals a clearer structure. The Elena cluster shows slightly higher and more stable agreement across all dimensions (relevance: $\rho = 0.52$, validity: $\rho = 0.41$, completeness: $\rho = 0.33$), suggesting a coherent evaluative strategy. In contrast, the Leo cluster exhibits strong agreement only for relevance ($\rho = 0.59$), while agreement drops substantially for validity and completeness, indicating a narrower evaluation strategy focused primarily on ranking relevance rather than multi-criteria assessment.

To assess whether the constructed personas reproduce expert-like evaluation behavior, we prompted each persona (instantiated via GPT-4o-mini; see Methods) to rate the same explanations previously assessed by experts, and compared persona ratings with aggregated expert scores. Persona construction used qualitative feedback, whereas validation used quantitative ratings, although both were based on the same explanation set.
As also shown in Figure~\ref{fig:persona-discovery}c, persona ratings were more positively correlated with expert ratings across all three criteria than across participants. Elena showed positive alignment across relevance ($\rho = 0.56$), completeness ($\rho = 0.41$), and validity ($\rho = 0.53$), whereas Leo showed the strongest alignment for relevance ($\rho = 0.77$) followed by validity ($\rho = 0.53$) and completeness ($\rho = 0.47$).

These results suggest that the inferred personas reflect structured patterns in how experts evaluate explanation quality, even under limited sampling conditions.

\subsection*{Perspective explanations improve ratings across evaluation criteria}
To evaluate whether persona-conditioned explanations improve expert assessment, we conducted a larger persona evaluation study with 22 biomedical experts (68.2\% life sciences, 18.2\% hybrid computational biology, 13.6\% CS/AI with biomedical experience; demographics in Supplementary Table~\ref{tab:supp-userstudy}). Each participant was assigned to a persona via a three-question profiling instrument (Figure \ref{fig:persona-profiles}c) derived from discriminative characteristics identified in the formative expert feedback (see Methods). Participants evaluated ten drug repurposing hypotheses, each time comparing a perspective explanation with a general-purpose explanation on validity, relevance, and completeness; presentation order was randomized. REx was chosen as the general-purpose baseline by virtue of being the state-of-the-art for knowledge graph-based explanation generation~\cite{nunesRewarding2025}; no alternative general-purpose method currently achieves comparable explanation quality in this setting, making it a strong comparator against which to measure the value of adaptation. 

\begin{figure}[H]
    \centering
    \includegraphics[width=1.01\linewidth]{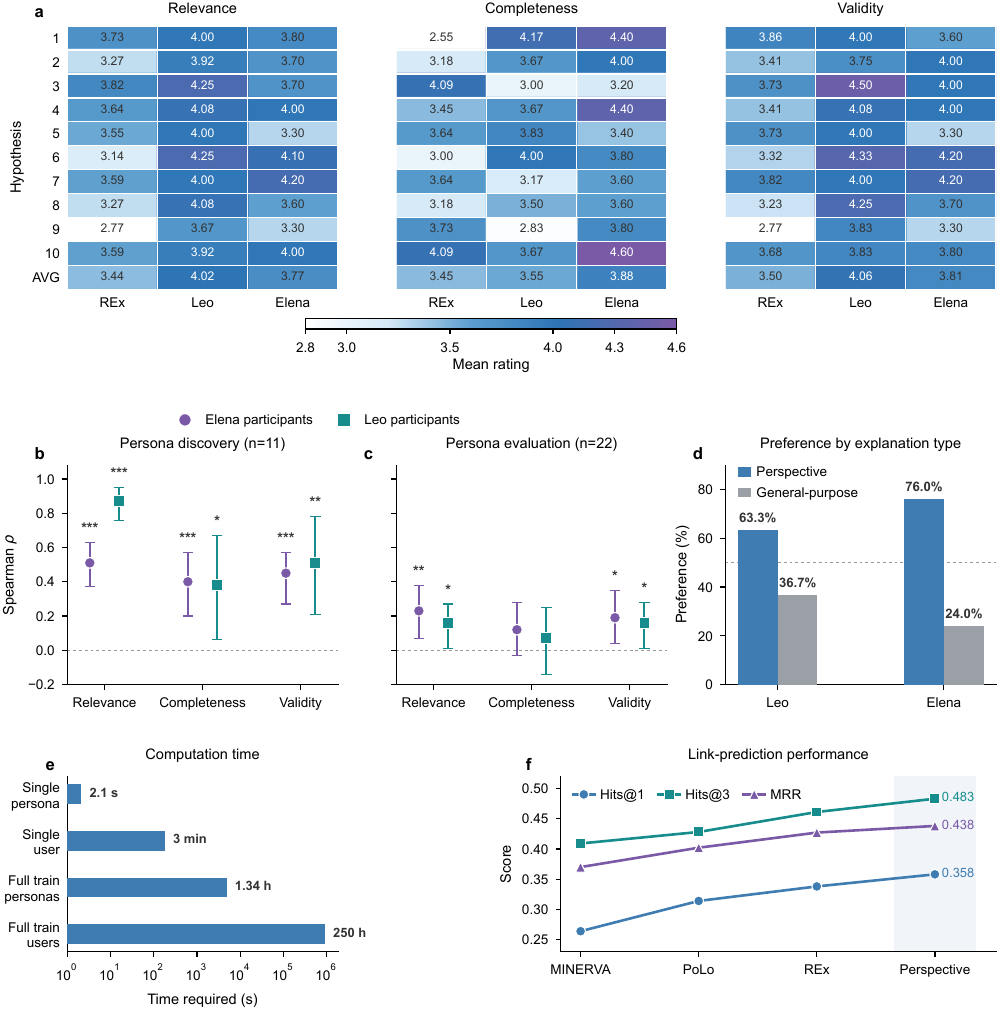}
\caption{Perspective explanation evaluation. \textbf{a},  Average ratings across evaluation criteria for general-purpose (REx) and persona-conditioned (Leo, Elena) explanations per hypothesis, with REx ratings reported as the shared baseline. \textbf{b}, Spearman correlation between persona ratings and their constituent participants' ratings for drug repurposing in the persona creation study.  \textbf{c}, Spearman correlation between persona ratings and their assigned participants' ratings for drug repurposing in the persona evaluation study. \textbf{d}, Participant preferences between perspective and general-purpose explanations. \textbf{e}, Estimated feedback collection time under different supervision conditions; persona-conditioned training reduces expert time by two orders of magnitude.  
\textbf{f}, Link prediction performance on the drug repurposing task; perspective explanations exceed state-of-the-art methods across all metrics. Asterisks in \textbf{b} and \textbf{c} denote significance: $^{*}p<.05$, $^{**}p<.01$, $^{***}p<.001$.}
\label{fig:evaluation}
\end{figure}

The profiling instrument assigned 12 participants to Leo and 10 to Elena. Figure \ref{fig:evaluation}a reveals higher ratings for perspective explanations over the general-purpose baseline across all three evaluation criteria and both personas. For relevance, Leo's explanations achieved the highest average score of $4.02$, compared to general-purpose explanations at $3.44$. Elena's explanations also outperformed general-purpose ones with $3.77$. Validity showed similar patterns, with Leo averaging $4.06$ and Elena $3.81$ versus general-purpose at $3.50$. The completeness criterion showed Elena achieving the largest gain with $3.88$ compared to $3.45$ for general-purpose explanations, while Leo also showed improvement at $3.55$. 

We further examined how persona benefits varied across the ten individual drug repurposing hypotheses (H1--H10, each corresponding to a specific drug-disease pair; Figure~\ref{fig:evaluation}a). Completeness ratings for some hypotheses showed substantial improvements with persona alignment: for instance, H1 increased from $2.55$ (general-purpose) to $4.40$ (Elena), and H10 rose from $4.09$ to $4.60$. Participants' justifications illuminate these gains, with one noting that the general-purpose explanation for H1 ``is too complex, as it considers a lot of similar drugs...the links are sometimes confusing,'' while Elena's version was praised for being ``simpler and it is enough for understanding the hypothesis.'' However, not all hypotheses benefited equally: for H3, the general-purpose baseline remained competitive at $4.09$, exceeding both Leo ($3.00$) and Elena ($3.20$). Validity improvements were particularly strong for H6 ($3.32$ to $4.33$ with Leo) and H3 ($3.73$ to $4.50$ with Elena).

Statistical analysis using Wilcoxon signed-rank tests confirmed these improvements. Overall, perspective explanations significantly outperformed general-purpose explanations for relevance ($W = 3170.5, p < .001, r = 0.076$) and validity ($W = 2637.0, p < .001, r = 0.083$), though completeness showed no significant difference ($W = 4518.5, p = .107$). When analyzed separately by persona, Elena-aligned explanations showed significant improvements across all three metrics (relevance: $W = 295.5, p < .001$; completeness: $W = 728.0, p = .016$; validity: $W = 291.5, p < .001$), while Leo-aligned explanations improved significantly only for relevance ($W = 1459.5, p = .010$) and validity ($W = 1097.0, p = .024$), with no improvement in completeness ($W = 1592.5, p = .890$). While statistically significant, the effect sizes were small ($r < 0.10$), indicating modest but consistent improvements.

A linear mixed-effects model with crossed random intercepts for participants and hypotheses, which accounts for the repeated-measures structure, confirms these effects: perspective explanations score significantly higher on relevance ($\beta = 0.47$, $p<.001$) and validity ($\beta = 0.45$, $p<.001$), with completeness directional but not significant ($\beta = 0.25$, $p = .057$); the improvement holds for both persona groups (Elena $\beta = 0.57$, Leo $\beta = 0.24$; both $p<.01$; Supplementary Table~\ref{tab:supp-mixed}).

The improvements across metrics were also strongly interrelated. Spearman correlation analysis revealed that improvements in relevance correlated with validity ($\rho = 0.699, p < .001$) and completeness ($\rho = 0.548, p < .001$). Validity and completeness improvements were moderately correlated ($\rho = 0.383, p < .001$), suggesting that perspective explanations provided coherent benefits across multiple dimensions of explanation quality.

\subsection*{Perspective explanations generalize to new users, suggesting more broadly shared epistemic preferences}
To understand whether personas capture more broadly shared epistemic perspectives, we evaluated persona–user correlations in an out-of-sample setting using participants from the explanation evaluation study. This allows us to assess whether participants’ ratings of explanations align with those of their questionnaire-assigned persona beyond the population used for persona construction. This further enables a closer examination of Leo’s coherence, given its small formative base but substantial assignment in the study ($n=12$).

We prompted GPT-4o-mini with each persona narrative to evaluate the same 10 drug repurposing explanations assessed by the participants. For each hypothesis, the persona rated both the perspective and general-purpose explanations on relevance, completeness, and validity using the same scales and instructions as human participants. We then computed Spearman correlations between persona ratings and aggregated participant ratings (Figure \ref{fig:evaluation}b and \ref{fig:evaluation}c).

Across the two studies, persona–expert correlations were lower in the independent evaluation study than in the discovery study, but the pattern remains meaningful. In the discovery study ($n = 11$), persona alignment was strong for relevance and validity: Elena participants correlated with Elena at  $\rho = 0.51$ and  $\rho = 0.45$, while Leo participants correlated with Leo at $\rho=0.87$ and  $\rho = 0.51$, respectively. In the independent evaluation study ($n = 22$), these effects were smaller but remained consistently positive and significant for both personas: relevance generalized for Elena ($\rho = 0.23$) and Leo ($\rho = 0.16$), and validity likewise generalized for Elena ($\rho = 0.19$) and Leo ($\rho = 0.16$). This suggests that the personas capture stable differences in how participants assess whether an explanation is relevant to the hypothesis and whether it is scientifically valid. By contrast, completeness was less robust: although it showed positive correlations in the discovery study, it dropped to weak, non-significant correlations in the evaluation study for both personas. Overall, the results support the claim that persona-conditioned explanations generalize best for relevance and validity, while completeness appears to be a more variable and harder-to-model dimension of explanation quality.

\subsection*{Experts strongly prefer perspective explanations}
Participants showed a clear preference for perspective explanations (Figure~\ref{fig:evaluation}d). Leo and Elena users chose perspective options 63.3\% (binomial test, $p = .005$) and 76.0\% ($p < .001$) of the time, respectively. 
An analysis of participants’ free-text justifications 
revealed systematic persona differences. Leo participants cited general-purpose explanations as \textit{``too complex''} and \textit{``overwhelming''}, preferring explanations that were \textit{``concise while retaining essential components''} and excluded irrelevant information. Elena participants tolerated complexity when justified, valuing \textit{``a complete overview''} without interpretive overload. A recurring paradox emerged: general-purpose explanations were \textit{``overly detailed''} but  \textit{``failing to provide necessary information''}. Leo resolved this through conciseness and Elena through selective depth.
This preference is not explained by simplification alone. On the drug repurposing task, persona-conditioned explanations use paths of the same length as the general-purpose baseline (three hops in every retained path) yet select substantially different paths: in 65\% (Elena) and 50\% (Leo) of drug-disease explanations they share no path with REx (Supplementary Table~\ref{tab:supp-explprops}). They contain modestly fewer paths (Elena 1.83 versus 2.11 per pair; Leo 1.66 versus 2.16), consistent with both personas' stated preference for parsimony, so the expert preference reflects re-selected, perspective-aligned evidence rather than shorter or compressed explanations.

\begin{figure*}[tb!]
    \centering
    \includegraphics[width=0.80\linewidth]{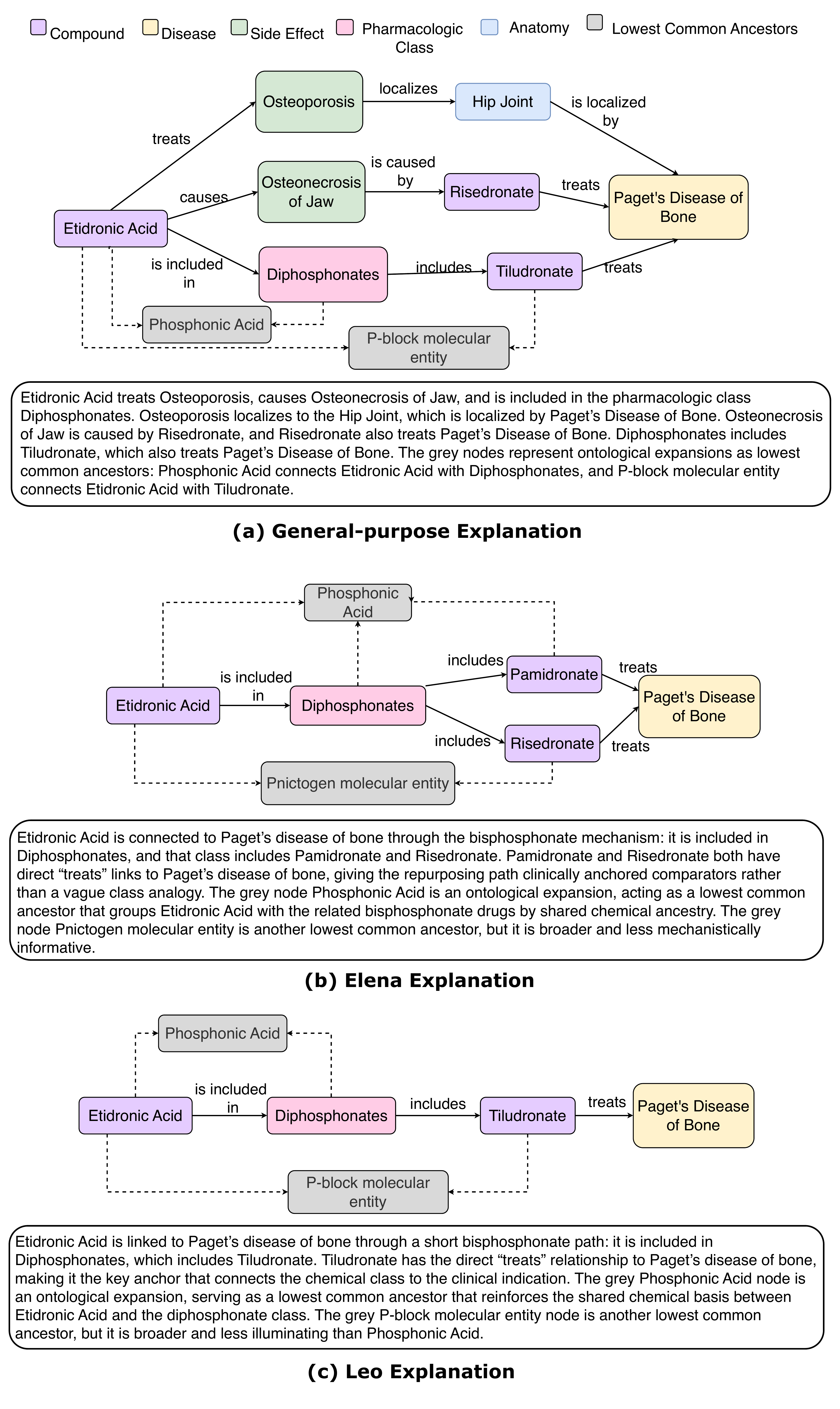}
        \caption{Example explanations for the hypothesis \textit{Etidronic acid treats Paget's disease of bone}: \textbf{a}, general-purpose (REx); \textbf{b}, the Elena perspective (mechanistic rigour); \textbf{c}, the Leo perspective (focused clarity).}
    \label{fig:example-3perspectives}
\end{figure*}

For instance, given the hypothesis \textit{Etidronic acid treats Paget's disease of bone}, the general-purpose explanation (Figure~\ref{fig:example-3perspectives}a) pairs the drug-class mechanism with peripheral evidence such as a side effect and an anatomical association. The Elena perspective (Figure~\ref{fig:example-3perspectives}b) retains only the mechanistic route, and the Leo perspective (Figure~\ref{fig:example-3perspectives}c) reduces the same mechanism to a single, directly relevant class route. The three explanations share the same path length but differ in which evidence each selects and how its verbalization frames it.

\subsection*{Persona-conditioned rewards preserve predictive quality and enable scalable training}
Beyond improving explanatory alignment, perspective-conditioned explanations should also preserve predictive utility. We therefore compared link prediction performance on the drug repurposing task between our perspective explanation model and state-of-the-art approaches, MINERVA~\cite{das2017go}, PoLo~\cite{liu2021neural}, and REx~\cite{nunesRewarding2025} (Figure~\ref{fig:evaluation}f). Perspective explanations achieved the highest scores across all metrics ($Hits@1: 0.358$, $Hits@3: 0.483$, $MRR: 0.438$), outperforming both MINERVA ($MRR: 0.370$) and the general-purpose REx baseline ($MRR: 0.427$). These results indicate that persona-conditioned rewards guide the agent toward higher-quality paths without sacrificing fidelity to the underlying prediction task.

Beyond predictive quality, persona-conditioned training enables scalable adaptation by removing the dependency on direct expert feedback for reinforcement learning. During training, personas scored approximately 5,000 candidate explanations before reward stabilization (see Methods). Replacing these evaluations with direct expert feedback would be impractical at this scale: as illustrated in Figure~\ref{fig:evaluation}e, a full human-feedback training loop would require approximately 250 hours (10.4 days) of expert time, whereas persona-conditioned training completes in 1.34 hours, a 187$\times$ speedup. At the level of individual evaluations, a single persona call scoring one explanation takes 2.1 seconds, compared to an average of 3 minutes for expert feedback, an 86$\times$ reduction. During training, candidate explanations are scored in batches of up to 20 per call, so the effective cost per explanation is lower than that of a single call.

\section*{Discussion}
This work reveals a fundamental challenge in explainable AI for scientific discovery: while experts share core requirements for explanatory quality, they diverge in how they interpret specific explanation features. We propose that rather than reflecting noise, this divergence derives from distinct epistemic perspectives that cut across professional backgrounds and tasks. We therefore introduce perspective explanations as a mechanism for aligning explainable model outputs with these differing interpretive frames. 

To investigate this, we adopt knowledge graph path-based explanations because they simultaneously produce predictions and structured, relational explanations characteristic of scientific inquiry. This representation also enables a systematic analysis of how experts evaluate explanations, as paths expose multiple dimensions of explanatory quality that align with explanation quality criteria (i.e., relevance, validity, and completeness). Our experiments across different drug development tasks provide evidence that expert explanation preference is structured rather than idiosyncratic. We cluster participants based on their qualitative evaluations of explanations to identify patterns of epistemic preference, showing that expert judgment can organize around distinct interpretive perspectives.

We capture these patterns through epistemic personas modeled on expert feedback and use their ratings of explanations as reward signals in a reinforcement learning framework over the knowledge graph. Persona ratings across explanation quality criteria capture characteristic patterns of expert evaluation, with positive out-of-sample alignment for relevance and validity, indicating that personas generalize beyond the individuals used to construct them. This enables explanation generation to be conditioned on inferred perspectives without requiring large-scale expert supervision, reducing the estimated expert time required for training by two orders of magnitude and addressing the practical constraint that expert feedback does not scale in scientific settings.

Our experiments show that even a small number of empirically-derived personas is sufficient to drive the generation of perspective-conditioned explanations. Crucially, they exceed state-of-the-art systems in drug repurposing predictive tasks, suggesting that epistemic alignment and predictive fidelity are complementary rather than competing objectives. Our results also speak to the longstanding debate in XAI over potential trade-offs between explainability and predictive performance~\cite{gunning2021darpa,london2019artificial,erasmus2021interpretability,rudin2019stop}: in our setting, perspective alignment improves how experts evaluate explanations without sacrificing predictive performance. Importantly, experts consistently prefer perspective-conditioned explanations over general-purpose baselines. An analysis of qualitative justifications reveals that participants described general-purpose explanations as simultaneously overloaded and incomplete, a contradiction that persona conditioning resolves through different strategies depending on the epistemic perspective. 

These findings should be interpreted within the scope of our empirical study. The epistemic perspectives were derived from a relatively small formative cohort with an uneven distribution across the identified groups, and therefore should not be considered an exhaustive account of expert reasoning. Although perspective alignment persisted in an independent cohort, generalization was stronger for relevance and validity than for completeness. Moreover, our evaluation is situated in drug discovery, and further studies are needed to determine whether comparable structures of epistemic variation emerge across other scientific domains and expert communities.

Our results support a view of explainability as an epistemically plural process, in which the effectiveness of an explanation depends partly on the perspective through which it is evaluated. This shift motivates several new research directions, including methods for discovering richer and more fine-grained persona spaces as well as their validation across larger and more diverse expert populations, mechanisms for real-time adaptation of explanatory content, structure, and modality to align with different epistemic perspectives, and frameworks for integrating multiple perspectives within collaborative scientific settings. In this sense, perspective explanations are not merely a tool for improving explainability, but can become a foundation for building AI systems that can adapt to the diverse ways in which experts interpret and evaluate scientific evidence.

\section*{Methods}

\subsection*{Problem Formulation}
We define a knowledge graph as $\mathcal{G} = (\mathcal{E}, \mathcal{R}, F)$, where $\mathcal{E}$ is the set of entities, $\mathcal{R}$ is the set of relations, and $F \subseteq \mathcal{E} \times \mathcal{R} \times \mathcal{E}$ is the set of triples denoted as $(s,r,o)$ for subject, relation, and object.
Building on~\cite{keas2018systematizing,akujuobi2024link,nunesRewarding2025} we establish the following definitions:
\begin{definition}[Hypothesis] A hypothesis is a predicted link between a subject $s$ and an object $o$, $h = (s,r,o)$, e.g., (Fenofibrate, treats, Coronary Artery Disease).
\end{definition}

\begin{definition}[Explanation]
Given a hypothesis $h = (s,r,o)$, an \emph{explanation} is a set of paths
\[
\mathcal{X}(h) = \{ p_1, \dots, p_m \}
\]
such that each path $p_j$ in $\mathcal{G}$ is selected by maximizing some criteria $\sigma_i$ for explainability:
\[
p_j \in \arg\max_{p: s \to o} \; f(\sigma_1(p),...,\sigma_{K}(p))
\]
where $f: \mathbb{R}^K \rightarrow \mathbb{R}$ is an aggregation function over $K$ explainability criteria $\{\sigma_i\}_{i=1}^{K}$.
\end{definition}

For instance, MINERVA maximizes \textit{fidelity} (whether the explanatory path successfully connects $s$ and $o$)~\cite{das2017go}, whereas REx maximizes both fidelity and \textit{relevance} (how informative an explanatory path is, favoring paths that traverse specific, detailed entities over generic, highly connected ones)~\cite{nunesRewarding2025}. 

However, there may be multiple paths connecting the subject and object of a hypothesis triple that fit the criteria of scientific explainability. Critically, it is established that experts even within the same field present a diversity of epistemic perspectives, and current explanation generation approaches are unable to capture them. Here, we propose \textit{perspective explanations}, i.e., explanations that are conditioned on particular epistemic perspectives.

\begin{definition}[Perspective-shaped explanation]
Given a hypothesis $h = (s,r,o)$ and an epistemic perspective $\theta$, a perspective-shaped explanation or \emph{perspective explanation} is a set of paths
\[
\mathcal{X}(h \mid \theta) = \{ p_1, \dots, p_m \}
\]
such that each path $p_j$ is selected by jointly maximizing fidelity, relevance, and epistemic alignment:
\[
p_j \in \arg\max_{p: s \to o} \; f\big(\alpha(p),\beta(p),\gamma(p,\theta)\big),
\]
where $\alpha$ is the fidelity of a path, $\beta$ is its relevance, $\gamma(p,\theta)$ is the epistemic perspective score of $p$ under $\theta$, and $f$ is a multi-objective aggregation function.
\end{definition}

\subsection*{Creating epistemic personas}
We conducted a persona creation study with 11 biomedical experts assessing knowledge graph–based explanations across drug repurposing and drug–target interaction tasks (demographics in supplementary Table \ref{tab:supp-formative}). Some participants evaluated both tasks, yielding 15 written responses in total (9 DR, 6 DTI). The purpose of this study was to identify systematic differences in expert interpretive preferences and to collect feedback necessary for persona construction.

Participants evaluated explanations generated from Hetionet~\cite{himmelstein2017systematic}, a heterogeneous biomedical knowledge graph integrating multiple data sources across drugs, genes, diseases, and biological processes, containing $\sim$45{,}000 entities, $\sim$4.5M triples, and 48 relation types after pre-processing. We presented ten hypotheses for two different tasks, drug repurposing (DR) and drug-target interaction (DTI) prediction. In each case, four explanations per hypothesis generated by different KG-based systems that prioritize different aspects were presented: MINERVA~\cite{das2017go} (fidelity); PoLo~\cite{liu2021neural} (fidelity and logic rules),  REx~\cite{nunesRewarding2025} (fidelity and relevance), and RExLight (an ablated version of REx without ontological enrichment). Participants included PhD students, researchers, and professors with backgrounds in life sciences ($n=4$), CS/AI with biomedical experience ($n=5$), and hybrid computational biology ($n=2$).

Participants rated explanations using a 5-point Likert scale along three dimensions derived from established frameworks of explanatory virtues in scientific theories~\cite{keas2018systematizing,nunes2026ESWC}: relevance, completeness, and validity.
We also collected extensive qualitative feedback for each explanation (see Supp. Materials) to inform the construction of epistemic personas.

\subsubsection*{Persona Generation}
Our methodology draws from Shin et al.~\cite{shin2024understanding}, who compared multiple approaches for persona generation and found that LLM-based summarization of clustered behavioral data yields the most coherent and actionable results. We adopt a three-stage process: semantic embedding of expert feedback, unsupervised clustering, and LLM-based narrative synthesis.

To mitigate the risks of persona essentialization noted above~\cite{cabrero2016critique}, we specifically design our personas as epistemic perspective proxies rather than fixed user representations. 
Our approach: (i) captures explanatory preferences from observed expert rationales, not demographics; (ii) maintains traceability to cluster-level evidence with transparency documentation; (iii) focuses on task-specific evaluative language, minimizing identity descriptors; and (iv) validates persona credibility against aggregated expert ratings before deployment.

\paragraph{Input Representation and Embedding}
The persona creation study drew on 125 expert statements describing preferences and judgments about explanations, providing rich qualitative feedback suitable for clustering.

Prior to embedding, all free-text comments were manually curated to ensure consistency and anonymity. We reformulated each statement into a standardized format using first-person declaratives such as \textit{``I prefer explanations that...''} or \textit{``I avoid explanations where...''}. This normalization preserved the subjective tone of the original feedback while enabling semantic alignment across responses. We removed references to explanation IDs or system names. Each curated response was aggregated into a single text block and encoded into a 768-dimensional dense vector using Sentence-BERT~\cite{reimers-2019-sentence-bert} (\texttt{all-mpnet-base-v2}~\cite{song2020mpnet}). Embeddings were L2-normalized.

\subsubsection*{Clustering and Profile Discovery}
User response embeddings were clustered using three strategies: centroid-based (K-Means), hierarchical (Agglomerative), and density-based (HDBSCAN)~\cite{jain2010data}. We evaluated cluster quality using Silhouette Score~\cite{rousseeuw1987silhouettes}, Davies--Bouldin Index~\cite{davies2009cluster}, Calinski--Harabasz Index~\cite{calinski1974dendrite}, and Inertia~\cite{thorndike1953belongs} across candidate solutions from $k=2$ to $k=5$. To validate the selected solution beyond internal metrics, we performed Gaussian mixture model selection (BIC and AIC), bootstrap cluster stability analysis (200 resampled datasets, measured via Jaccard similarity of co-assignment matrices), and subcluster checks within the majority cluster (splitting via K-Means ($k=2$) and applying HDBSCAN to test for density-based subgroups), complemented by per-sample silhouette analysis to verify individual assignment quality. Final clusters were selected based on convergence across metrics, algorithm agreement, and stability, balancing interpretability with internal consistency.

\paragraph{Narrative Generation via LLM}
To translate clusters into actionable personas, we prompted an instruction-tuned LLM (OpenAI o3-pro) with all user records belonging to a given cluster, including curated preference statements, background metadata\footnote{Each participant's background was classified into: \textit{Life Sciences}, \textit{Computer Science/AI}, \textit{Hybrid Computational Biology}, \textit{CS/AI-heavy with Biomedical Experience}, and \textit{Social Sciences/Legal (Other)}.}, and structured analytical instructions (prompt template in Supp. Materials \ref{lab1}). The prompt guided the model to identify explanatory themes by frequency, distinguishing core traits ($>$40\% of participants), secondary traits (25--40\%), and weak signals ($<$25\%); generate a naturalistic character profile; and produce an evidence summary documenting for each trait: (i) the number of participants who expressed it, (ii) the percentage of the cohort, and (iii) the disciplinary backgrounds of those participants.

\subsection*{Generating Perspective Explanations}
We extend a reinforcement learning framework~\cite{nunesRewarding2025} for finding explanatory paths on knowledge graphs and integrate persona-conditioned rewards to align generated explanations with expert epistemic perspectives. Unlike post-hoc explanation pipelines that enumerate candidate paths and rank or cluster them with auxiliary models~\cite{perdomo2026generating}, our approach optimizes explanation generation end-to-end by shaping the RL reward with persona-conditioned signals during training. The full system consists of three tightly coupled components: (i) RL-based path discovery, (ii) persona-conditioned evaluation, and (iii) curriculum-guided optimization. 

\subsubsection*{RL-based explanatory path discovery.}
We adopt and extend the reinforcement learning approach of~\cite{das2017go} for hypothesis validation on knowledge graphs. The environment is formulated as a deterministic partially observable Markov decision process. A state encodes the current entity and the hypothesis under evaluation:
\[
S = (e, s_h, o_h),
\]
where $e$ denotes the agent position and $(s_h, o_h)$ define the hypothesis subject and object. The agent observes only its current location and the subject entity, i.e., $O(S) = (e, s_h)$.

At each step, the agent selects either a neighboring edge $(e, r, e')$ or a STOP action, producing a path that connects entities in the knowledge graph. Transitions are deterministic, and a trajectory terminates when STOP is chosen or a terminal condition is reached. We compute explanatory paths using a policy network based on LSTMs, which encodes the interaction history and maps it to a distribution over valid actions.

To construct final explanations, we aggregate discovered paths and group them by metapath structure. Within each group, we retain the most informative path (as measured by relevance) and merge selected paths into a subgraph enriched with lowest common ancestors, yielding a coherent explanation graph $\mathcal{G}_h$.

\subsubsection*{Persona-conditioned reward modeling.}
To incorporate expert epistemic perspectives, we introduce a persona-conditioned reward function. Each persona represents a cluster of expert evaluation strategies derived from qualitative feedback.

At the end of each rollout, a selected persona evaluates the generated explanation path along three dimensions: validity ($v$), completeness ($c$), and relevance ($r$). The persona reward is defined as:
\[
R_{\text{persona}} = \frac{1}{3}(v + \hat{c} + r),
\]
where $\hat{c}$ rescales completeness so its ideal midpoint scores highest. The scores are normalized to $[0,1]$, and equal weighting is used for all dimensions.

\paragraph{LLM-based persona evaluation.}
Personas are operationalized using GPT-4o-mini, prompted with a structured persona narrative. Each knowledge graph path is linearized into a natural-language explanation describing entity–relation sequences and contextualized by the original drug–disease hypothesis. The model produces scalar evaluations for the three criteria. The choice of GPT-4o-mini balances reasoning capability with computational cost and can be substituted via configuration.

\paragraph{Curriculum-guided reinforcement learning.}
Since integrating persona evaluations into every rollout is computationally expensive, requiring one language model call for each produced path, we introduce a self-paced curriculum that gradually increases the selectivity of paths submitted to persona scoring. We adapt the fidelity and relevance reward scores proposed in~\cite{nunesRewarding2025} as selective mechanisms.

Fidelity ($\alpha$) ensures that a valid path connects the hypothesis entities:
\[
\alpha(p) =
\begin{cases}
1, & \text{if } e_T = o_h, \\
0, & \text{otherwise}.
\end{cases}
\]

The relevance ($\beta$) of a path quantifies how informative its underlying causal connections are for explaining a hypothesis. This is computed using the information content (IC), where less frequent (higher surprisal) entities and relations contribute more informative signals. 

The IC of a node $e \in \mathcal{E}$ measures the surprisal of that node appearing in a randomly sampled triple:
\[
\text{IC}(e) = -\log \frac{\deg(e)}{\max_{v \in \mathcal{V}_{\mathcal{G}}} \deg(v)}
\]
where $\text{deg}(e)$ is the total degree of entity $e$ in the knowledge graph and $\left(\max_{v \in \mathcal{V}_{\mathcal{G}}} \deg(v)\right)$ denotes the maximum degree over all nodes $v$ in the graph $\mathcal{G}$. To correct for potential biases arising from different granularities in node degree, we modify the IC computation by clustering edge types according to their similarity, computed via K-means clustering of OWL2vec*~entity embeddings~\cite{chen2021owl2vec}, with the number of clusters set to 10\% of the total node count.

Specifically, we compute path relevance as the average IC of its constituent edges, where each edge IC is defined as the mean IC of its connected entities. To mitigate biases arising from heterogeneous node granularity and uneven study coverage in knowledge graphs, the IC is further refined using a clustered variant that groups semantically similar entities into clusters and computes surprisal over the resulting clustered graph. This reduces distortions caused by redundant entity representations or research bias, providing a more robust estimate of path informativeness. Relevance is therefore given by

\[
\beta(p) = \frac{1}{|p|} \sum_{(u,r,v) \in p} \frac{IC(u) + IC(v)}{2}
\]

Only paths satisfying fidelity and a relevance threshold $\tau$ are considered for reward assignment:
\[
R(p, t) =
\begin{cases}
R_{\text{persona}}(p), & \text{if } \beta(p) > \tau_{\text{relevance}}(t), \\[2mm]
0.25, & \text{if } 0.5 \le \beta(p) \le \tau_{\text{relevance}}(t), \\[1mm]
0.10, & \beta(p) < 0.5.
\end{cases}
\]

We define a relevance threshold $\tau_{\text{relevance}}(t)$ that increases over training:
\[
\tau_{\text{relevance}}(t) =
\begin{cases}
0.50 & t < 60,\\
0.55 & 60 \le t < 80,\\
0.60 & 80 \le t < 100,\\
0.65 & \text{otherwise}.
\end{cases}
\]

This design ensures that high-quality paths are aligned with persona-specific epistemic judgments, while lower-quality paths still contribute to exploration through shaped rewards. The curriculum thus progressively shifts learning from coarse hypothesis connectivity toward fine-grained alignment with expert evaluation strategies, while significantly reducing LLM evaluation cost by restricting persona scoring to high-relevance candidates. At inference time, the threshold is fixed to the best-performing validation setting to ensure consistency. Training and policy-network hyperparameters otherwise follow the defaults of the underlying RL framework~\cite{nunesRewarding2025}. Persona scoring is performed sequentially, with up to 20 candidate explanations from the same hypothesis scored in a single LLM call. Averaged across five seeds, each persona-conditioned training run scored approximately 6,500 candidate explanations for Elena and 4,700 for Leo before reward stabilization.

\subsection*{Evaluation}
\subsubsection*{Epistemic personas}
To understand whether epistemic personas can capture expertise diversity and effectively serve as proxies for epistemic perspectives, we compared personas' ratings for the same 40 explanations per task (DR and DTI) with the original assessment by human experts. The automated scoring presented explanations from all four systems (MINERVA, PoLo, REx, and RExLight), mirroring how experts naturally compare alternatives. Persona ratings were compared against aggregated expert scores.

\subsubsection*{Perspective Explanations}
We recruited 22 participants with biomedical backgrounds, including PhD students, postdocs, and researchers from both academic and industry settings (68.2\% life sciences, 18.2\% hybrid computational biology, 13.6\% CS/AI with biomedical experience). Participants were primarily early-career researchers with strong molecular biology expertise (72.7\% Competent or Expert) but limited KG/AI specialization, reflecting our target population of domain experts evaluating scientific explanations (see demographics in supplementary Table \ref{tab:supp-userstudy}).

Each participant evaluated ten drug repurposing hypotheses, viewing two explanations per hypothesis (20 total evaluations): one standard REx (general-purpose) and one persona-conditioned (perspective). We chose REx as the general-purpose baseline to ensure a controlled comparison under identical architecture and training conditions. Both general-purpose and perspective explanations were generated by the models that achieved the best validation performance. Presentation order was randomized. 

\paragraph{Persona Assignment.}
Participants were assigned to personas via three single-choice questions derived from each persona's expert feedback. These questions captured discriminative characteristics between Elena and Leo: preferences for mechanistic detail (detailed step-by-step vs.\ streamlined), explanation complexity (direct mechanistic links vs.\ high-level ontological context), and predicate specificity (specific relations like ``binds'' vs.\ broader terms like ``causes''). Each question offered two response options, each aligned to a persona. A detailed supplementary table \ref{tab:supp-persona-mapping} maps user quotes to questions. Participants were assigned to the persona matching the majority of their responses (Elena: $n=10$; Leo: $n=12$). Both question and response order were randomized.

\paragraph{Explanation Evaluation.}
For each explanation, participants rated validity, relevance, and completeness, then selected their preferred explanation with free-text justification. Graph-based explanations were accompanied by textual verbalizations generated via GPT-4o-mini (Figure~\ref{fig:example-3perspectives}). The verbalization system (prompt template in Supp. Materials \ref{lab3}) produced paragraph-style descriptions capturing pathways and relationships without generic summarization; for perspective explanations, verbalization incorporated the persona profile to reflect that persona's interpretive priorities.

\paragraph{Predictive Performance.} We also compared link prediction performance on the drug repurposing task between our perspective explanation model and state-of-the-art approaches \newline~\cite{das2017go,liu2021neural,nunesRewarding2025} to ensure predictive power was not sacrificed. Drug repurposing is cast as link prediction over Hetionet's $(\textit{Compound}, \textit{treats}, \textit{Disease})$ triples, with 483 hypotheses used for training, 121 for validation, and 151 held out for testing.
All models were evaluated on the same test set using default parameters and standard ranking metrics: Hits@1, Hits@3, and Mean Reciprocal Rank (MRR). These metrics assess how well each model ranks the correct target entity among all candidates for a given drug-disease query.

\section*{Acknowledgements}
This work was supported by FCT through the fellowship \\ https://doi.org/10.54499/2023.00653.BD, and the LASIGE Research Unit, \\ ref. UID/00408/2025. It was also partially supported by the KATY project, which has received funding from the European Union’s Horizon 2020 research and innovation program under grant agreement No. 101017453, and by the CancerScan project, which received funding from the European Union’s Horizon Europe Research and Innovation Action (EIC Pathfinder Open) under grant agreement No. 101186829. Views and opinions expressed are, however, those of the author(s) only and do not necessarily reflect those of the European Union or the European Innovation Council and SMEs Executive Agency. Neither the European Union nor the granting authority can be held responsible for them.

\section*{Ethics approval and consent to participate}
Ethics committee approval is not required for this type of study at our institution. Both studies consisted of anonymous online evaluation tasks completed by adult domain experts, involved no intervention, and collected no personal identifiers or sensitive data; participants were identified only by random codes. Before taking part, all participants were informed of the purpose of the studies, that participation was voluntary and could be withdrawn at any time without consequences, and that anonymized results could be used in scientific publications. All participants gave informed consent.

\section*{Code availability}
All code required for reproducing all results in this study is publicly available at \url{https://github.com/liseda-lab/Perspective_XAI}. 

\section*{Data availability}
The Hetionet knowledge graph is publicly available at \url{https://het.io}. All remaining data required for reproducibility is available at\url{https://github.com/liseda-lab/Perspective_XAI}.

\section*{Competing interests} The authors declare no competing interests.

\section*{Author contributions}
Contributions here
S.N. and C.P. designed the method. S.N. developed the method, conducted the user studies, performed analyses, and wrote the first draft of the manuscript. S.N., T.G., and C.P. designed the study. T.G. and C.P. supervised the research. T.G. and C.P. reviewed and revised the manuscript. All authors approved the final version.








\begin{appendices}



\section{Supplementary Data}

\begin{table}[htb!]
\centering
\footnotesize
\caption{Persona creation study demographics, expertise, and domain knowledge ($n = 11$).}
\label{tab:supp-formative}
\begin{tabular}{p{0.25\linewidth} p{0.65\linewidth}}
\toprule
 & \textbf{Study (n = 11)} \\
\midrule
\textbf{Age Range} &
25--34: 9; 35--49: 2 \\
\midrule
\makecell[l]{\textbf{Primary} \\ \textbf{Background}} &
Life and Health Sciences: 4 \newline
CS/AI-heavy w/ Biomedical Experience: 5 \newline
Hybrid Computational Biology: 2 \\
\midrule
\makecell[l]{\textbf{Domain} \\ \textbf{Knowledge}} &
Knowledge Graphs: 4 Experts, 2 Competent, 3 Novice, 2 No Knowledge \newline
AI Systems: 1 Expert, 7 Competent, 1 Novice, 2 No Knowledge \newline
Molecular Biology: 2 Experts, 8 Competent, 1 No Knowledge \\
\bottomrule
\end{tabular}
\end{table}

\begin{figure}[htb!]
{\sffamily \textbf{a}}\\
\vspace{0pt}
\begin{center}
\includegraphics[width=.9\linewidth]{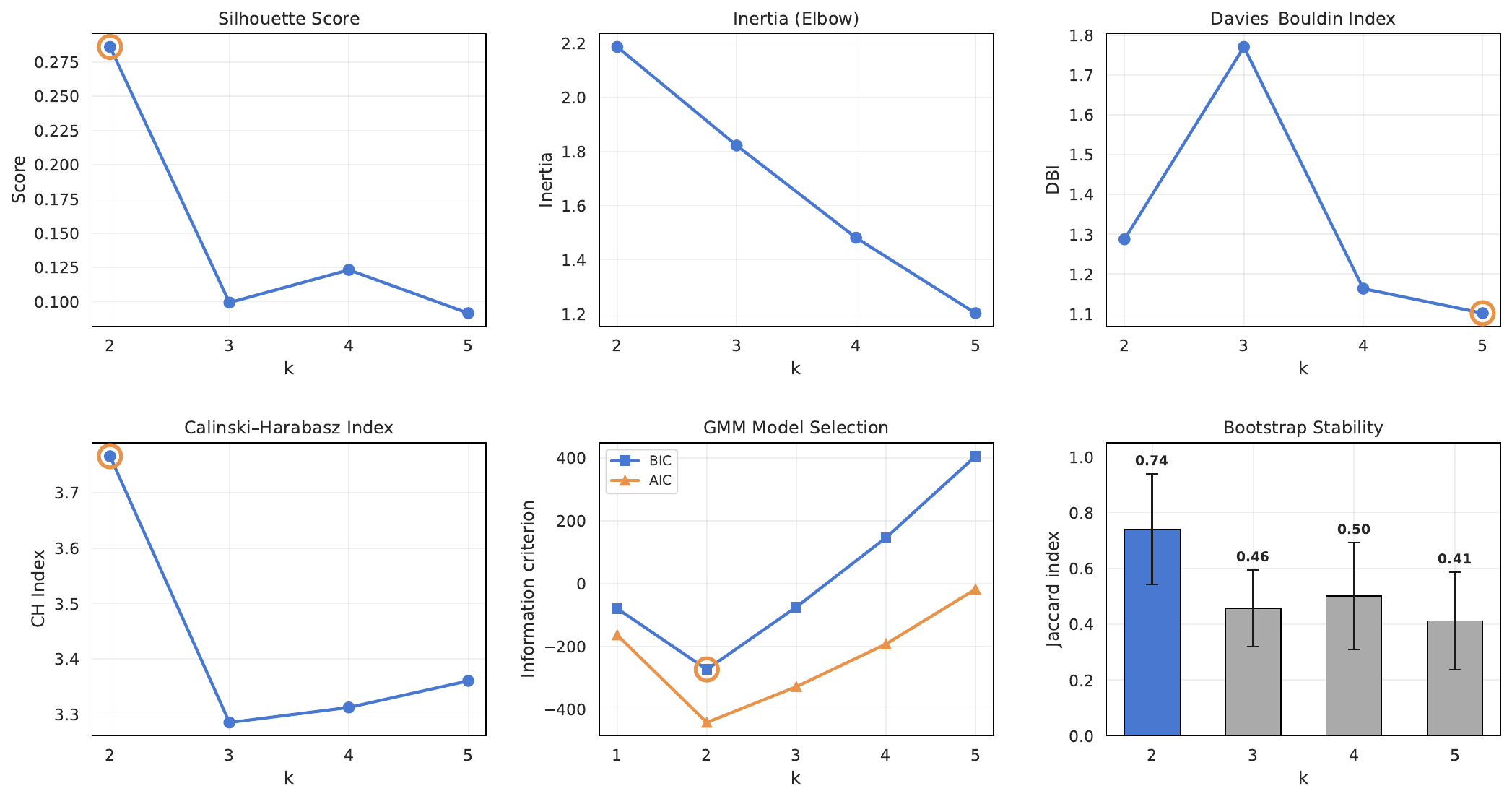}
\end{center}

{\sffamily \textbf{b}}\\
\vspace{0pt}
\begin{center}
\includegraphics[width=.9\linewidth]{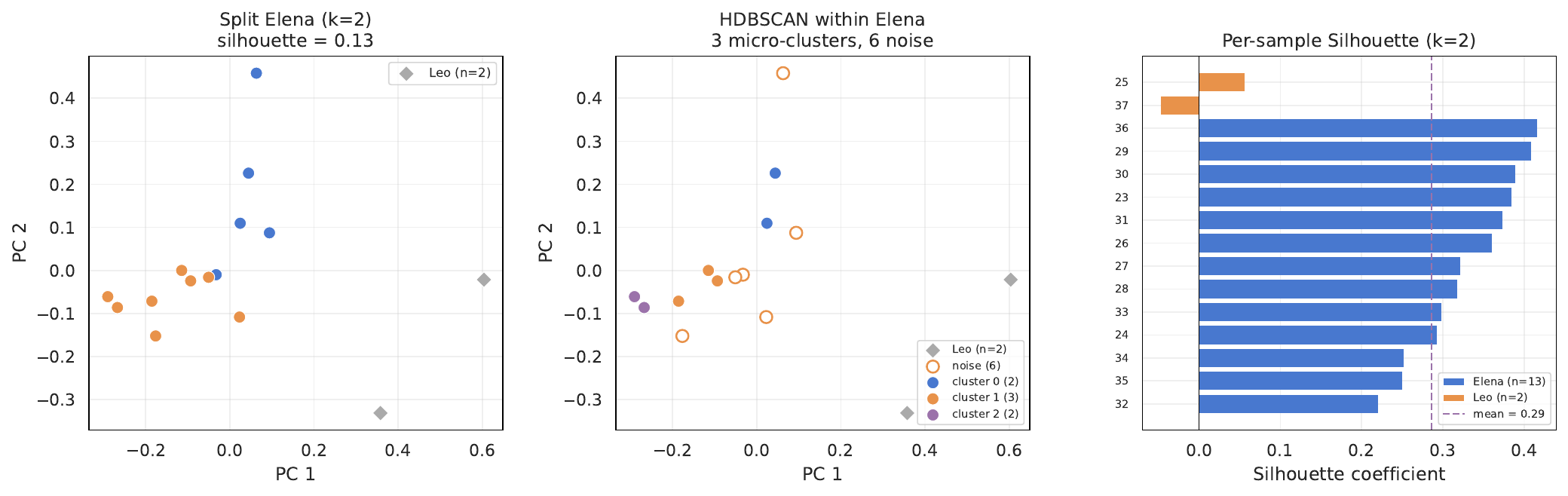}
\end{center}

\caption{Clustering validation. \textbf{a}, Candidate solutions ($k=2$–$5$): Silhouette Score, Inertia, Davies--Bouldin Index, Calinski--Harabasz Index, Gaussian mixture model selection (BIC and AIC), and bootstrap cluster stability (Jaccard index, 200 resamples). Orange rings indicate the best $k$ for each metric. \textbf{b}, Subcluster analysis within the majority cluster (Elena, $n=13$): K-Means split ($k=2$, silhouette = 0.13), HDBSCAN (3 micro-clusters, 6 noise points), and per-sample silhouette coefficients for the base $k=2$ solution.}
\label{fig:clustering-validation}
\end{figure}

\begin{table}[htb!]
\centering
\footnotesize
\caption{Persona evaluation study participant demographics, background categories, main expertise areas, and domain knowledge ($n = 22$).}
\label{tab:supp-userstudy}
\begin{tabular}{p{0.25\linewidth} p{0.30\linewidth} p{0.30\linewidth}}
\toprule
\textbf{Category} & \textbf{Distribution} & \textbf{} \\
\midrule

\textbf{Age Range} & \multicolumn{2}{p{0.62\linewidth}}{20--24: 2; 25--29: 12; 30--34: 4; 40--44: 3; 45--54: 1} \\
\midrule

\textbf{Persona Assigned} & \multicolumn{2}{p{0.62\linewidth}}{Leo: 12; Elena: 10} \\
\midrule

\textbf{Primary Background} &
\multicolumn{2}{p{0.62\linewidth}}{
Life and Health Sciences: 15 (68.2\%) \newline
CS/AI-heavy with Biomedical Experience: 3 (13.6\%) \newline
Hybrid Computational Biology: 4 (18.2\%)
} \\
\midrule

\multirow{9}{*}{\textbf{Main Expertise Area}}
 & Biochemistry: 2 & Animal Cell Technology: 1 \\
 & Microbiology: 2 & Data Analytics: 1 \\
 & Computer Science: 2 & Biochemistry and Biotechnology: 1 \\
 & Biotechnology: 2 & Biochemistry, Computer Science: 1 \\
 & Biomedical Engineering: 1 & Bioinformatics, Computer Science: 1 \\
 & Medicine: 1 & Health and Bioinformatics: 1 \\
 & Pharmacy: 1 & Forensic Science: 1 \\
 & Bioinformatics: 1 & Veterinary Medicine: 1 \\
 & Marine Biology: 1 & Biological Engineering: 1 \\
\midrule

\textbf{Domain Knowledge} & \multicolumn{2}{p{0.62\linewidth}}{} \\
\quad Knowledge Graphs & \multicolumn{2}{p{0.62\linewidth}}{Expert: 5; Competent: 2; Novice: 11; No Knowledge: 4} \\
\quad AI Systems       & \multicolumn{2}{p{0.62\linewidth}}{Expert: 1; Competent: 5; Novice: 12; No Knowledge: 4} \\
\quad Molecular Biology & \multicolumn{2}{p{0.62\linewidth}}{Expert: 5; Competent: 11; Novice: 6} \\
\bottomrule
\end{tabular}
\end{table}

\begin{table}[htb!]
\centering
\footnotesize
\caption{Linear mixed-effects model of explanation ratings ($n=22$; 1{,}320 observations) with crossed random intercepts for participant and hypothesis, fit by REML. Coefficients give the perspective-minus-general difference on the 1--5 scale.}
\label{tab:supp-mixed}
\begin{tabular}{lcccc}
\toprule
\textbf{Contrast} & \textbf{$\beta$} & \textbf{SE} & \textbf{95\% CI} & \textbf{$p$} \\
\midrule
Relevance              & $+0.468$ & 0.082 & $[+0.31,+0.63]$ & $<.001$ \\
Completeness$^\dagger$ & $+0.245$ & 0.129 & $[-0.01,+0.50]$ & $.057$  \\
Validity               & $+0.450$ & 0.085 & $[+0.28,+0.62]$ & $<.001$ \\
\midrule
Overall                & $+0.388$ & 0.059 & $[+0.27,+0.50]$ & $<.001$ \\
\quad Elena users      & $+0.570$ & 0.081 & $[+0.41,+0.73]$ & $<.001$ \\
\quad Leo users        & $+0.236$ & 0.084 & $[+0.07,+0.40]$ & $.005$  \\
\bottomrule
\end{tabular}

{\footnotesize $^\dagger$Convergent completeness coding (higher $=$ better).}
\end{table}

\begin{table}[htb!]
\centering
\footnotesize
\caption{Path properties of general-purpose (REx) and persona-conditioned explanations on the drug repurposing task.}
\label{tab:supp-explprops}
\begin{tabular}{lccccc}
\toprule
\textbf{Comparison} & \textbf{Pairs} & \makecell{\textbf{Persona}\\\textbf{paths/pair}} & \makecell{\textbf{REx}\\\textbf{paths/pair}} & \makecell{\textbf{Path length}\\\textbf{(hops)}} & \makecell{\textbf{Pairs sharing}\\\textbf{no path}} \\
\midrule
Elena vs REx & 46 & 1.83 & 2.11 & 3.0 & 65\% \\
Leo vs REx   & 44 & 1.66 & 2.16 & 3.0 & 50\% \\
\bottomrule
\end{tabular}
\end{table}

\begin{table}[htb!]
\centering
\caption{Mapping of profiling questions to persona characteristics based on expert feedback analysis.}
\label{tab:supp-persona-mapping}
\begin{tabular}{p{1.5cm}|p{5.5cm}|p{5.5cm}}
\toprule
\textbf{Question} & \textbf{Elena} & \textbf{Leo} \\
\midrule
\textbf{Q1:} Mechanistic Detail &
\textit{``A detailed step-by-step explanation''} \newline
\textbf{Supporting statements:} \newline
- ``combine therapeutic, genetic, and adverse-effect details for depth'' \newline
- ``combine multiple mechanistic or genetic details to reinforce direct drug-disease links'' &

\textit{``A streamlined explanation with fewer steps''} \newline
\textbf{Supporting statements:} \newline
- ``offer multiple paths for completeness while limiting their number to avoid excessive complexity'' \newline
- ``concise explanations that link drugs by shared treatment lines'' \\
\midrule

\textbf{Q2:} Explanation Complexity &
\textit{``I do not need high-level concepts, I am happy with seeing only a detailed explanation''} \newline
\textbf{Supporting statements:} \newline
- ``explicitly state direct, clinically established drug-disease links'' \newline
- ``avoid generic class or side-effect links, deliver a clear therapeutic rationale'' &

\textit{``I prefer to see high-level concepts that frame why things belong together''} \newline
\textbf{Supporting statements:} \newline
- ``use ontological expansion to clarify connections between claims'' \newline
- ``leverage ontology for non-obvious or indirect relations'' \\
\midrule

\textbf{Q3:} Decision-making Context &
\textit{``Explanations should be based on specific relations such as `inhibits' or `binds'\,''} \newline
\textbf{Supporting statements:} \newline
- ``prefer `binds' to `causes' for drug-target interactions'' \newline
- ``eliminate vague predicates like `causes,' `includes,' or `participates'\,'' &

\textit{``Explanations can also include less specific relations such as `treats' or `causes'\,''} \newline
\textbf{Supporting statements:} \newline
- ``relations\ldots already covered by `causes' or `treats'\,'' (accepting them) \newline
- ``ontological expansion specifically to indirect relations'' \\
\bottomrule
\end{tabular}
\end{table}

\clearpage
\section{Prompts}
\subsection{Persona Synthesis Prompt}
\label{lab1}
The following prompt was used with OpenAI's \texttt{o3-pro} model to synthesize natural language personas from clustered user feedback.

\begin{lstlisting}[basicstyle=\ttfamily\footnotesize,breaklines=true]
Below are individual user feedback records from a group
who share similar preferences. Each user has provided
their background and detailed preferences.

Core Requirements:
- Analyze frequency patterns: Count how many users mention
  each theme
- Weight by user count: Prioritize themes mentioned by
  multiple users
- Flag weak signals: Mark any trait mentioned by <25% of
  users as potentially unreliable
- Create coherent character narratives: Write profiles as
  if describing a specific individual
- Validate with evidence: Support all persona traits with
  concrete user counts and percentages

Key Instruction:
Write the Profile section as if describing a real person's
preferences and behaviors. Never mention "users", "this
group", or research language within the Profile. Save all
analytical language for the Evidence Summary.

Persona Structure:
Name: Descriptive name
Tagline: One sentence capturing the strongest supported
  theme (40%+ users)
Profile:
- Lead paragraph: Core traits (40%+ support)
- Secondary paragraph: Additional values (25-40%)
- "Also values..." section: Traits in 15-25% range
- Background note: Brief mention of role diversity
Evidence Summary:
- Core traits (40%+): List with counts and percentages
- Secondary traits (25-40%): List with counts
- Weak signals (<25%): List or omit
- Background composition: Summary
\end{lstlisting}

\subsection{RL Reward Prompt}
\label{lab2}
At the end of each RL rollout, the active persona rates each path according to the following prompt (example shown for Elena):

\begin{lstlisting}[basicstyle=\ttfamily\footnotesize,breaklines=true]
You are evaluating explanation paths from the perspective
of the following persona:
"Name: Elena
Tagline: Trusts nothing but rock-solid mechanistic logic.
Profile: [Full persona narrative inserted here]"

Score EACH path individually on three criteria:
1. Scientific Validity (V): 1-5. Scientific correctness,
   plausibility, and coherence.
2. Completeness (C): 1-5, where 3 is ideal. 1 = too
   simple, 5 = too complex.
3. Relevance (R): 1-5. Usefulness for understanding
   why the prediction matters.

Output strictly in JSON format as an array of objects:
[{"id": <path id>, "validity": <int>,
  "completeness": <int>, "relevance": <int>}]
\end{lstlisting}

\subsection{Graph Verbalization Prompts}
\label{lab3}
Three verbalization detail levels were implemented (brief, standard, comprehensive), with and without persona conditioning. Below is the standard persona-conditioned prompt:

\begin{lstlisting}[basicstyle=\ttfamily\footnotesize,breaklines=true]
Adopt this persona completely:
[PERSONA_TEXT_INSERTED_HERE]

Write a natural, flowing description of the connections
in this drug repurposing explanation. The graph shows
paths from [DRUG] to [DISEASE].
Cover the key relationships in 4-6 COMPLETE sentences.
Write as connected prose, not as bullet points or lists.

CRITICAL REQUIREMENTS:
- Every sentence MUST reference specific path elements
- Do NOT write concluding or summary statements
- Focus ONLY on describing the specific connections
- Embody this persona's priorities, standards, and
  interests
- Filter the description through this persona's lens
- Match this persona's preferred depth, focus, and style
\end{lstlisting}

\end{appendices}


\bibliography{sn-bibliography}

\end{document}